\documentclass{article}

\usepackage{arxiv}
 \usepackage{amsmath} 
\usepackage[utf8]{inputenc} % allow utf-8 input
\usepackage[T1]{fontenc}    % use 8-bit T1 fonts
\usepackage{hyperref}       % hyperlinks
\usepackage{url}            % simple URL typesetting
\usepackage{booktabs}       % professional-quality tables
\usepackage{amsfonts}       % blackboard math symbols
\usepackage{nicefrac}       % compact symbols for 1/2, etc.
\usepackage{microtype}      % microtypography
\usepackage{lipsum}
\usepackage{graphicx}
\graphicspath{ {./images/} }
\usepackage{multirow}

\usepackage{subcaption}

\title{HuLLMI: Human vs. LLM Identification with Explainability}

\author{
 Prathamesh Dinesh Joshi \\
  Indian Institute of Information  \\
  Technology Design and Manufacturing,\\
  Kurnool, India 518008 \\
  \texttt{121cs0072@iiitk.ac.in} \\
   \And
Sahil Pocker \\
  Vizuara AI Labs\\
  Bangalore, India\\
  \texttt{sahilpocker98@vizuara.com} \\
  \And
 Raj Abhijit Dandekar \\
  Vizuara AI Labs\\
  Bangalore, India\\
  \texttt{raj@vizuara.com} \\
   \And
 Rajat Dandekar\\
   Vizuara AI Labs\\
  Bangalore, India\\
  \texttt{rajatdandekar@vizuara.com} \\
   \And
 Sreedath Panat\\
  Vizuara AI Labs\\
  Bangalore, India\\
  \texttt{sreedath@vizuara.com} \\
}

\begin{document}
\maketitle
\begin{abstract}
As LLMs become increasingly proficient at producing human-like responses, there has been a rise of academic and industrial pursuits dedicated to flagging a given piece of text as "human" or "AI". Most of these pursuits involve modern NLP detectors like T5-Sentinel and RoBERTa-Sentinel, without paying too much attention to issues of interpretability and explainability of these models. In our study, we provide a comprehensive analysis that shows that traditional ML models (Naive-Bayes, MLP, Random Forests, XGBoost) perform as well as modern NLP detectors, in human vs AI text detection. We achieve this by implementing a robust testing procedure on diverse datasets, including curated corpora and real-world samples. Subsequently, by employing the explainable AI technique  LIME, we uncover parts of the input that contribute most to a model's prediction, providing insights into the detection process. Our study contributes to the growing need for developing production-level LLM detection tools, which can leverage a wide range of traditional as well as modern NLP detectors we propose. Finally, the LIME techniques we demonstrate also have the potential to equip these detection tools with interpretability analysis features, making them more reliable and trustworthy in various domains like education, healthcare, and media.
\end{abstract}

% keywords can be removed

\section{Introduction}

\subsection{Background and Significance}

Large Language Models (LLMs) are becoming increasingly adept at a wide range of tasks such as machine translation, text completion, story generation, proofreading, grading examinations, and logical reasoning. LLMs signify a pivotal moment in the history of Artificial Intelligence (AI). In the closed source category, GPT-4o by OpenAI represents a significant advancement, offering enhanced capabilities for various applications\cite{openai2024gpt4o}. On the open-source front, Meta's LLaMA 3.1 has emerged as a powerful contender, bridging the gap between open and closed-source models with its impressive performance across multiple benchmarks\cite{meta2024llama3.1}.

As LLMs produce increasingly human-like responses, the challenge of distinguishing between human-authored and AI-generated text has grown significantly \cite{scarfe2024real}. This phenomenon has far-reaching implications across various sectors:

\subsubsection{Education}
In education, the use of LLMs like ChatGPT has raised concerns about academic integrity, as these models can generate responses that are indistinguishable from those of students. The potential for AI to complete assignments and exams on behalf of students challenges the authenticity of educational assessments and necessitates new strategies to ensure genuine student learning\cite{christian2023cnet}\cite{fridman2023health}.

\subsubsection{Journalism and Media}
The journalism industry is experiencing a transformation as AI-generated articles become more prevalent. This shift challenges the traditional roles of journalists and editors, as AI can rapidly produce content, potentially affecting the quality and reliability of news\cite{sison2023chatgpt}\cite{georgiou2024differentiating}.

\subsubsection{Cybersecurity}
In cybersecurity, LLMs have been termed "weapons of mass deception" due to their potential misuse in crafting sophisticated phishing attacks and other cyber threats. The ability of these models to generate convincing and contextually relevant text poses significant challenges for cybersecurity professionals\cite{jawahar2020automatic}\cite{ranade2024}. Additionally, the integration of generative AI into cybersecurity systems introduces new vulnerabilities, such as data poisoning and adversarial attacks, which can compromise system integrity and lead to unauthorized access\cite{nttdata2023}.

\subsubsection{Creative Industries}
Creative industries are grappling with the implications of LLMs as they are increasingly used to generate scripts, stories, and music. This raises complex questions about intellectual property rights and artistic authenticity, as AI-generated works blur the lines of traditional authorship\cite{guo2023close}\cite{salewski2023}\cite{xu2024}.

\subsubsection{Politics and Governance}
The ability of LLMs to generate persuasive and human-like text poses risks to democratic processes. AI-generated political content, including speeches and policy documents, could potentially manipulate public opinion and interfere with elections, highlighting the need for vigilance and regulatory frameworks\cite{chen2023}\cite{serrano2021digital}.

An important aspect of LLMs is their continual evolution; as they are being trained on increasingly diverse datasets and improving architectures, the complexity of differentiating human-generated text from machine-generated text becomes ever more challenging. The rapid advancements of models, including their ability to mimic human writing styles, introduce significant hurdles for even the most sophisticated detection techniques. This ongoing escalation underscores the critical need for the development of effective detection methodologies that can adapt alongside these evolving AI systems. Failing to keep pace with advancements in LLMs may result in increased misinformation and the undermining of trust in various domains where accurate information is vital \cite{georgiou2024differentiating, tan2023llm}. There is also a pressing need for effective and robust detection mechanisms to address potential misuse and to uphold the integrity of various domains where AI-generated text may infringe upon human creativity and intellectual property \cite{sison2023chatgpt, jawahar2020automatic}.

The proliferation of AI-generated misinformation presents numerous second-order issues, such as the erosion of public trust, the spread of harmful health misinformation, and the potential for fake datasets to contaminate machine learning models. Misinformation can lead to significant societal impacts, including polarization, public health risks, and economic instability \cite{fridman2023health, serrano2021digital, ecker2022psychological}. Moreover, the generation of fake datasets can undermine the reliability of AI systems, leading to biased or inaccurate outcomes \cite{arxiv2023fakenews}. 

This makes it clear that along with the huge amount of interest, investment, and acceleration in developing new and better LLMs, there should be an equal effort in developing effective methods that can flag test outputs as LLM-based. 

\subsection{Current Approaches to LLM Text Detection}

Detection methods can be broadly categorized into traditional machine learning techniques, advanced Natural Language Processing (NLP) methods, stylometric analysis, and hybrid approaches that combine elements of these methods. Traditional approaches include Logistic Regression and Support Vector Machines \cite{hu2023radar}, while advanced NLP methods encompass models like RoBERTa \cite{raffel2020exploring} and encoder-decoder-based detectors such as GPT-sentinel and RADAR, which utilize models like t5-small and Vicuna-7B respectively \cite{chiang2023vicuna}.

\subsubsection{Traditional Machine Learning Techniques}

Traditional machine learning techniques utilize algorithms such as Support Vector Machines (SVM), Naive Bayes, Decision Trees, and Random Forests. These methods rely on feature engineering, using statistical and linguistic features like TF-IDF, n-grams, and part-of-speech tags to differentiate between human and machine-generated text. They are valued for their interpretability and have been successfully applied to detect text generated by models like GPT-2 and GPT-3 \cite{hu2023radar}.

\subsubsection{Advanced NLP Methods}

Advanced NLP methods leverage deep learning models such as RoBERTa \cite{raffel2020exploring}, and encoder-decoder based detectors such as GPT-sentinel and RADAR, which utilize models like t5-small and Vicuna-7B respectively \cite{chiang2023vicuna}. These models are fine-tuned on datasets containing both human-written and LLM-generated text, enabling them to implicitly capture subtle textual distinctions. They often employ supervised learning and have shown strong performance in detection tasks \cite{yang2023detecting}.

\subsubsection{Stylometric Analysis}

Stylometric analysis examines unique writing patterns and linguistic features characteristic of AI-generated text. It uses features such as lexical diversity, syntactic patterns, and readability scores to distinguish AI-generated content from human-written text. While useful, its effectiveness may be limited due to the evolving sophistication of LLMs \cite{kumarage2023stylometric}.

\subsubsection{Hybrid Approaches}

Hybrid approaches integrate elements of traditional machine learning, NLP, and stylometry. These methods combine the strengths of multiple approaches to improve detection accuracy and robustness. For example, combining stylometric analysis with deep learning models has shown promise in enhancing detection capabilities \cite{zhang2024hybrid}.

Although these models have been developed, there is a lack of comprehensive studies investigating the role of major machine learning models in detecting LLM-generated text across diverse real-world scenarios \cite{gokaslan2019openwebtext}.

\subsubsection{Challenges in Detection Methods}

Despite significant advancements, several challenges persist in accurately distinguishing AI-generated text from human-written content. One major challenge is the rapid evolution of LLMs, which continually improve their ability to mimic human writing styles. This evolution requires detection algorithms to be constantly updated and adapted to keep pace with new models and techniques\cite{chen2023can}.

Research highlights the difficulty in maintaining detection accuracy across various domains and languages. Many detection models are trained on specific datasets, limiting their generalizability to new or unseen data\cite{yang2023detecting}. This issue is compounded by the diversity of LLM outputs, which can vary significantly depending on input prompts and contextual usage\cite{zhang2024llm}.

Detection methods also face challenges from adversarial attacks. Techniques like paraphrasing and style transfer can alter AI-generated text to evade detection, underscoring the need for more resilient and adaptive strategies\cite{decoding2024}\cite{kirchenbauer2023watermarking}. Additionally, the reliance on supervised learning approaches is problematic due to the scarcity of labeled datasets that accurately represent the wide range of AI-generated content\cite{zhang2024hybrid}.

To address these challenges, researchers are exploring innovative approaches such as zero-shot and few-shot learning to improve detection accuracy without extensive retraining\cite{su2023zeroshot}. Hybrid models that combine multiple detection techniques are also gaining traction as a way to enhance robustness and adaptability\cite{zhang2024hybrid}.

The ongoing advancements in LLMs and their widespread application across various domains highlight the pressing need for robust detection methodologies. As AI-generated content becomes increasingly sophisticated, the challenge of accurately distinguishing it from human-written text intensifies. Our study focuses on evaluating the performance of traditional ML models alongside modern NLP detectors, emphasizing the importance of interpretability and explainability in these systems. This approach aims to provide insights into the detection process, setting the stage for a deeper exploration of the methodologies that can enhance the reliability and trustworthiness of AI detection tools\cite{decoding2024}.

\subsection{Research Gaps and Objectives}

Despite advancements in AI-generated text detection, significant gaps remain in the scientific literature. One major gap is the lack of comprehensive comparisons between traditional machine learning (ML) models and modern natural language processing (NLP) approaches for detecting LLM-generated content\cite{radford2018improving}. While modern NLP models like T5-Sentinel and RoBERTa-Sentinel are widely used, their effectiveness relative to traditional models such as Naive Bayes, MLP, Random Forests, and XGBoost remains underexplored\cite{sarang2023naive}.

Another gap is the limited evaluation of detection methods on diverse, real-world datasets. Most studies rely on standard training-testing corpora, which may not capture the variability and complexity of real-world text\cite{zhang2024llm}. This limitation highlights the need for testing procedures that incorporate both curated corpora and real-world samples to assess the generalizability and robustness of detection models.

Additionally, there is insufficient focus on the interpretability and explainability of detection models. As AI systems become more complex, understanding how these models make decisions is crucial for their adoption in sensitive domains such as education, healthcare, and media\cite{hochreiter1997long}\cite{chen2023can}. Techniques like Local Interpretable Model-agnostic Explanations (LIME) offer potential solutions, but their application in the context of AI-generated text detection is still limited\cite{datacamp2023}.

Our study aims to address these gaps by evaluating the performance of traditional ML models alongside modern NLP detectors, emphasizing the importance of interpretability and explainability in these systems. By focusing on these objectives, our research contributes to the development of production-level LLM detection tools that leverage a wide range of detectors. Furthermore, the interpretability analysis features demonstrated through LIME techniques have the potential to make these tools more reliable and trustworthy in various domains, including education, healthcare, and media. This comprehensive approach aims to bridge existing gaps and enhance the effectiveness of AI detection methodologies\cite{decoding2024}.

\section{Methodology}

\subsection{Dataset generation}
We have two datasets that are used predominantly in this study: (a) AI-generated dataset and (b) Human-generated dataset. To obtain the AI-generated dataset, we used the OpenGPT Text Dataset curated by \cite{chen2023gpt}. This dataset consists of 29.395 textual samples which were generated by \texttt{gpt-3.5-turbo}. The authors took the \texttt{OpenWebText} data \cite{gokaslan2019openwebtext} as reference, and asked \texttt{gpt-3.5-turbo} to \texttt{Rephrase the text, paragraph by paragraph}. We get the human-generated dataset from \texttt{OpenWebText}\cite{gokaslan2019openwebtext}, which is a widely used publicly available resource that has aggregated massive amounts of data from Reddit threads, which have a minimum of three votes. This data is based on the original \texttt{WebText} data corpus described by \cite{radford2018improving} and is thus highly reliable. The dataset was compiled in 2019, and hence, it is highly unlikely that is generated by a Large Language Models (LLMs). Both datasets are balanced, containing an equal number of AI-generated and human-generated texts to ensure unbiased model training and evaluation. \newline

\subsection{Data preprocessing}
Subsequently, we implemented a robust data preprocessing pipeline. Initially, duplicate data and newline characters were removed from the dataset. The text data is tokenized using the \texttt{Tokenizer} class from \texttt{Keras}. The tokenizer is configured so that it can handle a maximum of 5000 words. Out-of-vocabulary words are replaced with a special token \texttt{00V}. The dataset is subsequently split into training and testing, with 80\% of the data being used for training, and the remaining 20\% being used for testing.

\subsection{Modeling}

\begin{table}[h]
    \centering
    \begin{tabular}{|p{2cm}|p{1.5cm}|p{1.7cm}|p{1.2cm}|p{2.0cm}|p{2.0cm}| p{2.0cm}|}
        \hline
        \textbf{Hyper-Parameters} & \textbf{Naive Bayes} & \textbf{Logistic \ Regression} & \textbf{Random Forests} & \textbf{XGBoost} & \textbf{MLP} & LSTM\\
        \hline
        Epoch & NA & 100 & NA & NA(iterative) & 200 & 20\\
        \hline
        Batch Size & NA & NA & NA & NA(Boosting Iterations) & 200 & 256\\
        \hline
        Learning Rate & NA & NA & NA & 0.3 & constant & 0.001\\
        \hline
        Weight Decay & NA & NA & NA & 1 & 0.0001 & Not Explicitly Set\\
        \hline
        Optimizer & NA & LBFGS & NA & GBTREE & ADAM & ADAM\\
        \hline
        Loss Function & Logloss & Logloss & Gini & Logloss & Logloss & Binary Cross Entropy\\
        \hline
        Scheduler & NA & NA & NA & NA & NA & NA\\
        \hline
        Data Set & \multicolumn{6}{|c|}{OpenGPTText-Final} \\
        \hline
    \end{tabular}
    \caption{Hyper-Parameters for all the traditional ML models we used in our study (NA Stands for Not Applicable)}
    \label{tab:hyperparameters}
\end{table}

\begin{table}[h]
    \centering
    \begin{tabular}{|c|c|c|}
        \hline
        \textbf{Hyper-Parameters} & \textbf{RoBERTa-Sentinel} & \textbf{T5-Sentinel} \\
        \hline
        Epoch & 15 & 5 \\
        \hline
        Batch Size & 512 & 512 \\
        \hline
        Learning Rate & $1 \times 10^{-4}$ & $5 \times 10^{-4}$ \\
        \hline
        Weight Decay & $1 \times 10^{-3}$ & $1 \times 10^{-3}$ \\
        \hline
        Optimizer & AdamW & AdamW \\
        \hline
        Loss Function & Cross entropy & Cross entropy \\
        \hline
        Scheduler & Cosine annealing & Cosine annealing \\
        \hline
        Data Set & OpenGPTText-Final & OpenGPTText-Final \\
        \hline
    \end{tabular}
    \caption{Hyper-Parameters for RoBERTa-Sentinel and T5-Sentinel}
    \label{tab:hyperparameters}
\end{table}

\subsubsection{Traditional ML Models}

After the data was preprocessed, we transferred the data to our ML modeling pipeline. We used the following five traditional ML methods to train our LLM detector:

\begin{itemize}
    \item Naive Bayes
    \item Logistic Regression
    \item Random Forests
    \item XGBoost
    \item Multi-Layer Perceptron (MLP)
\end{itemize}

These models were selected to represent a diverse range of traditional approaches. Naive Bayes and Logistic Regression provide baseline performance for probabilistic and linear models, respectively. Naive Bayes is particularly effective for text classification tasks due to its simplicity and efficiency in handling high-dimensional data\cite{Sarang2023}. Logistic Regression, on the other hand, is valued for its interpretability and ability to model binary outcomes\cite{hosmer2013applied}.

Random Forests and XGBoost represent ensemble methods known for their robustness and performance across various tasks. Random Forests use multiple decision trees to improve predictive accuracy and control over-fitting by averaging the results\cite{breiman2001random}. XGBoost, a gradient boosting framework, is recognized for its speed and performance, often outperforming other algorithms in structured data competitions\cite{chen2016xgboost}.

The Multi-Layer Perceptron (MLP) serves as a simple feedforward neural network baseline. MLPs are capable of capturing non-linear relationships in data, making them suitable for complex classification tasks\cite{bishop1995neural}.

For each of the above models, we implemented three steps in the modeling pipeline:

\begin{itemize}
    \item \textbf{Count Vectorizer}: Each sentence in a document is converted into a vector of word counts (Bag of Words). This approach provides a simple representation of text data, capturing the frequency of words without considering their order\cite{harris1954distributional}.
    
    \item \textbf{TF-IDF Transformer}: The word counts are adjusted to also give importance to the frequency with which the word appears in a document. Term Frequency-Inverse Document Frequency (TF-IDF) is a statistical measure that evaluates the importance of a word in a document relative to a corpus. It helps in reducing the impact of common words that are less informative\cite{salton1988term}. This transformer also includes normalization, which scales the TF-IDF scores so that the model inputs are normalized, improving model convergence and performance.
    
    \item \textbf{Model Training}: After the vectorization and TF-IDF transformation steps, the normalized word vectors are passed on to the model. The model then trains on the input data (LLM and Human) and the output label (0 for Human and 1 for LLM).
\end{itemize}

We noticed that adding the TF-IDF Transformation step led to a significant increase in the testing accuracy of the different models. This improvement underscores the effectiveness of TF-IDF in enhancing the representation of text data by emphasizing informative words over common ones. The model hyperparameters, optimizers, and training epochs are detailed in Table 1.

\subsubsection{RoBERTa-Sentinel and T5-Sentinel}
Apart from the traditional ML models, we also studied the RoBERTa-Sentinel and T5-Sentinel models proposed by \cite{chen2023gpt}. The model configurations we used for this were the same as those used by \cite{chen2023gpt} and have been added in Table 2.

\subsection{Testing}

To rigorously evaluate the models and ensure the absence of data leakage, we curated a custom dataset consisting of 25 human-written samples and 25 AI-generated samples. This custom dataset was created to provide an additional testing layer that neither overlaps with the training nor the initial testing datasets, thereby mitigating any potential data leakage bias that could artificially inflate model performance. Testing on a completely unseen dataset helps ensure that the models' performance is reflective of their generalization ability. To ensure that the custom dataset is as diverse as possible, we have collected 5 Human Written Samples in each of the below 5 domains: 

\begin{itemize}
    \item English Literature: The passages were selected from the book \texttt{101 Essays That Will Change the Way You Think by Brianna Wiest}. This book offers a range of writing styles and thought-provoking content, making it ideal for capturing the diversity found in English literature.
    \item Recipes by MasterChefs: Extracts were chosen from Recipes - \texttt{MasterChef (Indian TV Show)} on \texttt{Scribd.com}. Five recipes were carefully selected from this source, providing structured and instructional text by renowned MasterChefs.
    \item Tweets from 2019: To capture concise and informal communication styles typical of social media, we searched for "famous 2019 tweets" using Google and randomly chose five tweets from the search results.
    \item IMDb User Reviews from 2018: We visited the \texttt{official IMDb website} and identified the top 5 movies of 2018. From these, We selected one featured review for each movie, ensuring a variety of user-generated content that offers a mix of opinions, descriptions, and informal language.
    \item Quora User Messages Related to IIT Entrance Examination Preparation: For educational and question-response style texts,We performed a Google search for "IIT JEE exam preparation questions" and chose samples from answers posted before 2019. These messages and answers provide a variety of informative content related to educational queries.
    
\end{itemize}

After getting human-generated data from all these sources, we passed this data to GPT 4o-mini and prompted the LLM as follows: \texttt{please summarize this one into 3 lines keeping the context as it is}. Subsequently, we opened a new chat and then prompted the LLM along with the summarized version given by the previous prompt as follows \texttt{Now elaborate on this topic with around 600 to 750 words keeping the context in mind}. This prompt is better than simple rephrasing since it makes sure that the LLM output is not biased towards the human output. We followed this approach for the English Literature, IMDB user reviews, and Quora user messages to construct a paragraph answer in an independent manner. For the Masterchef Recipes dataset and the 2019 Tweets, we encountered shorter sentence lengths. For these cases, we used the similar prompting as above but asked the LLM to keep the output sentence length similar to the input sentence. With this, we have datasets ranging from shorter text to longer text, and we evaluate the model performance on this diverse length of data.

Once the human-generated and LLM-generated custom testing data is obtained, we test the 6 traditional ML models and the RoBERTa-Sentinel, T5-Sentinel on this dataset. The results have been shown in Section 3 of this study.  

 \subsection{LIME Analysis}

Local Interpretable Model-agnostic Explanations (LIME) is a powerful technique used to understand and interpret the predictions made by complex machine learning models. It provides understanding into which features are most influential in a model's decision-making process, thereby enhancing the transparency and trustworthiness of AI systems\cite{ribeiro2016should}. LIME achieves this by perturbing the input data and observing changes in the model's predictions, allowing for the identification of specific features that significantly impact the outcome.

In our study, LIME was applied to evaluate feature importance in text classification models across various datasets, including the Open-GPTTest dataset and a custom test dataset. Using the "LimeTextExplainer" in Python, we identified the specific words that had the greatest impact on whether the text was classified as AI-generated or human-generated. This analysis was crucial for understanding the decision boundaries of our models and for identifying potential biases in their predictions.

To ensure the method's consistency across different models, we created a custom prediction function that processes tokenized inputs and returns predicted probabilities for each class, using a batch size of 32 for efficient computation. LIME works by slightly altering the input tokens and observing how these changes affect the model's predictions, enabling us to pinpoint the words most influential in the classification process.

We implemented LIME across six traditional models, including Naive Bayes, Logistic Regression, Random Forests, XGBoost, and Multi-Layer Perceptron (MLP), along with RoBERTa-Sentinel and T5-Sentinel. For each model, we plotted the probability of text being classified as human or AI-generated and highlighted the top 10 features contributing to each prediction. This approach provided clear explanations into the decision-making process of each model, making it easier to understand which features were most crucial in determining the final classification\cite{datacamp2023}.

Furthermore, LIME's ability to provide visual explanations for model predictions helps stakeholders in sensitive domains, such as healthcare and education, to better understand and trust AI systems. By revealing the inner workings of complex models, LIME facilitates more informed decision-making and supports the development of AI systems that are both effective and interpretable\cite{lime2023}.

The dataset and the entire code has been uploaded at this Github repository: 

\texttt{https://github.com/pdjoshi-30/HULLMI-HUMAN-VS.-LLM-IDENTIFICATION-WITH-EXPLAINABILITY}

\section{Results}
\subsection{
Introduction 
}
In this section, we discuss the performance of the Traditional ML Models on the OpenGPTText Final Dataset and compare it with RoBERTa-Sentinel and T5-Sentinel models. RoBERTa-Sentinel employs the RoBERTa model for extracting features, which are then classified by a multi-layer perceptron (MLP). This design uses the pre-trained RoBERTa model to capture detailed text features. Conversely, T5-Sentinel adapts the Text-to-Text Transfer Transformer (T5) model to frame classification as a sequence-to-sequence problem, generating output sequences that correspond directly to classification results. We will also discuss the performance of all 6  Traditional ML Models and Roberta-Sentinel, T5-Sentinel performance on the Custom Test Data. All these models have been tested on various Evaluation Metrics like Accuracy, F1-Score, FNR, FPR, ROC curve, and DET curves to ensure that we capture different aspects of the Model Performance.

\subsection{Evaluation Metrics}

In this study, our main focus is to detect whether the text is Human Written or AI-generated. It is a binary classification problem with Human written labeled as '0' and AI generated labeled as '1'. Several evaluation metrics are essential for a comprehensive assessment of model performance. To take into account the model performance as well as the probable imbalance in the dataset, we looked at the metrics shown in table \ref{tab:metrics}.

\begin{table}[h!]
\centering
\begin{tabular}{|c|c|p{7cm}|}
\hline
\textbf{Metric} & \textbf{Formula} & \textbf{Description} \\ \hline
Accuracy &
$\frac{TP + TN}{TP + TN + FP + FN}$ &
The proportion of correctly classified instances among the total instances. \\ \hline

\multirow{3}{*}{F1 Score} &
\multirow{3}{*}{$\frac{2 \cdot (Precision \cdot Recall)}{Precision + Recall}$} &
The harmonic mean of Precision and Recall, providing a balance between the two. \\
& & \\
& & \\ \hline

Precision &
$\frac{TP}{TP + FP}$ &
The proportion of true positives among all positive predictions. \\ \hline

Recall (True Positive Rate, TPR) &
$\frac{TP}{TP + FN}$ &
Measures how well the model identifies positive cases, ensuring important positives are not overlooked \\ \hline

False Positive Rate (FPR) &
$\frac{FP}{FP + TN}$ &
Measures the proportion of actual negatives incorrectly classified as positives, highlighting the frequency of false alarms\\ \hline

True Negative Rate (TNR) &
$\frac{TN}{TN + FP}$ &
Measures how effectively the model identifies negatives, minimizing false alarms.  \\ \hline

False Negative Rate (FNR) &
$\frac{FN}{FN + TP}$ &
Measures the proportion of actual positives missed by the model, which is critical in scenarios where failing to detect positives has severe consequences\\ \hline
\end{tabular}
\caption{Classification Metrics and Their Descriptions}
\label{tab:metrics}
\end{table}

Together, the metrics shown in table \ref{tab:metrics} provide a balanced view of model performance, addressing both the accuracy of positive and negative classifications and the implications of errors in various contexts.

In addition to these metrics, visual tools like the ROC curve display the True Positive Rate (TPR) against the False Positive Rate (FPR) across thresholds, visualizing the trade-off between sensitivity and specificity. The area under the ROC curve (AUC) measures overall performance, with higher values indicating better class distinction, and the  DET curve, plotted on a normal deviate scale, provides detailed information into error rate trade-offs, particularly useful for imbalanced classes and fine-tuning.

\begin{figure}[htbp]
    \centering
    \begin{subfigure}[b]{0.45\textwidth}
        \centering
        \includegraphics[width=\textwidth]{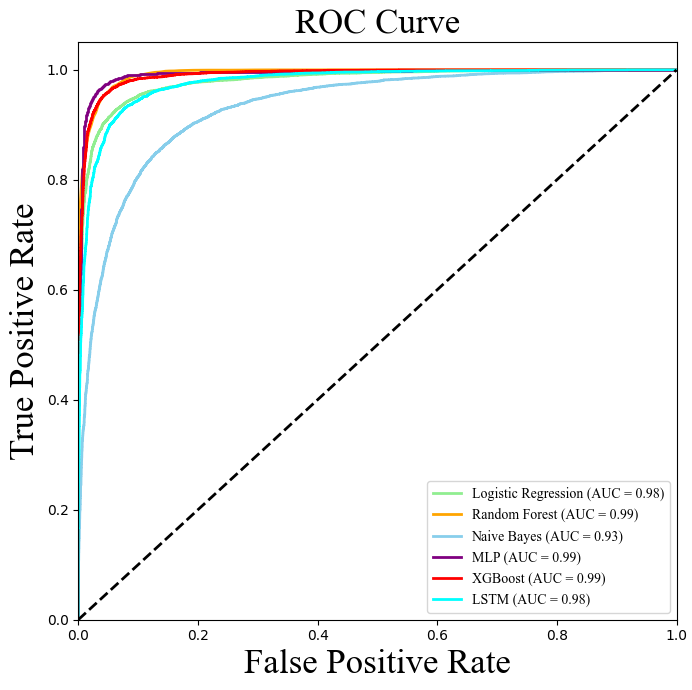}
        \subcaption*{(a)}
    \end{subfigure}
    \hfill
    \begin{subfigure}[b]{0.45\textwidth}
        \centering
        \includegraphics[width=\textwidth]{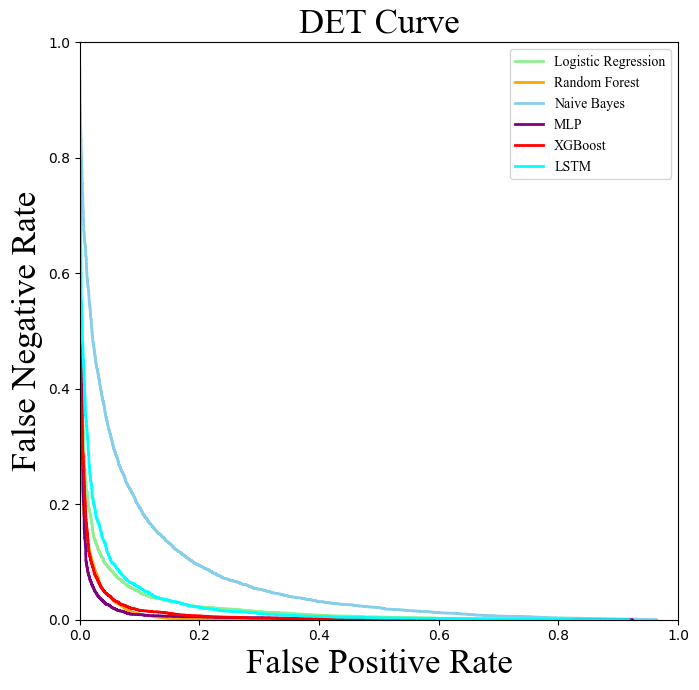}
        \subcaption*{(b)}
    \end{subfigure}

    \caption{The figure displays (a) Receiver Operating Characteristic (ROC) curves and (b) Detection Error Tradeoff (DET) curves for the various traditional ML models analyzed. The ROC curves illustrate the performance of each model in distinguishing between positive and negative classes. Notably, the Random Forest, XGBoost, and MLP models achieve the highest AUC values of 0.99, indicating excellent classification performance. Logistic Regression and LSTM follow closely with AUCs of 0.98, while Naive Bayes has a lower AUC of 0.93. The DET curves reveal that models with higher ROC AUCs, such as Random Forest, XGBoost, and MLP, tend to have lower DET rates, demonstrating that higher ROC values correspond to better overall performance. This indicates that these models are not only effective at distinguishing between classes but also balanced in minimizing both false positive and false negative rates.}
\end{figure}
\begin{figure}[htbp]
    \centering
    \begin{subfigure}[b]{0.45\textwidth}
        \centering
        \includegraphics[width=\textwidth]{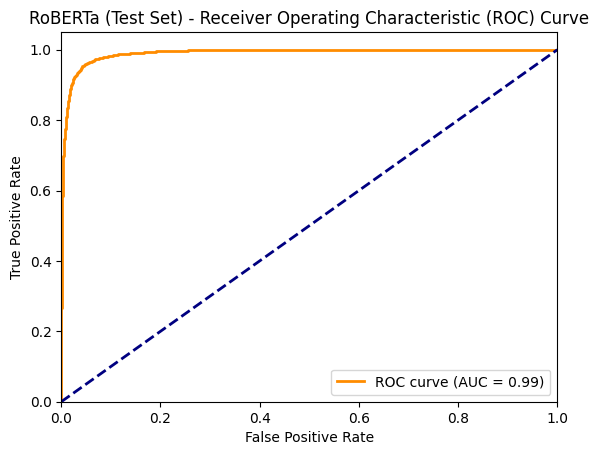}
        \subcaption*{(a)}
    \end{subfigure}
    \hfill
    \begin{subfigure}[b]{0.45\textwidth}
        \centering
        \includegraphics[width=\textwidth]{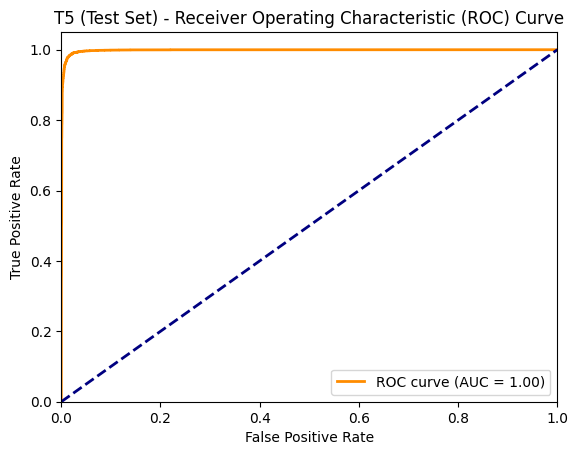}
        \subcaption*{(b)}
    \end{subfigure}

    \caption{Figure shows (a) Receiver Operating Characteristic (ROC) curve for Roberta and (b)  Receiver Operating Characteristic (ROC) curve for T5.}
\end{figure}

\subsection{
Analyzing Evaluation Metrics on Traditional ML Models
}
In the analysis of various traditional ML models applied to the OpenGPTText-Final dataset, it is observed that Naive Bayes has the lowest ROC value among all the models, as shown in Figure 2. This can be attributed to the assumption of feature independence, which might not hold true in complex datasets with large text samples and a significant number of samples\cite{Sarang2023}. Naive Bayes is known for its simplicity and generally performs well with smaller datasets. However, it struggles with larger, more complex datasets, which can lead to a low ROC value. As shown in Table 4, Naive Bayes has the highest False Negative Rate, indicating significant difficulty in correctly classifying positive instances.

On the other hand, models like Logistic Regression, Random Forest, MLP, and XGBoost exhibit stronger ROC values than Naive Bayes and outperform it by a considerable margin. LSTM, with an AUC of 0.98, can be attributed to its ability to capture long-term dependencies through its recurrent structure\cite{hochreiter1997lstm}, excelling in tasks with sequential patterns. MLP's neural network structure allows it to model non-linear relationships effectively, contributing to its high performance. Logistic Regression's strength lies in its robustness and interpretability, particularly for linearly separable data, while Random Forest's ensemble approach captures complex patterns by averaging the decisions of multiple trees, making it versatile across different datasets.

The top-performing traditional models in terms of F1-score are LSTM, XGBoost, and Logistic Regression. These models strike a balance between precision and recall, indicating their effectiveness in accurately identifying positive instances while minimizing false alarms. The F1-score highlights their reliability in handling binary classification tasks, particularly in distinguishing between human-written and AI-generated content, where both false positives and false negatives carry significant consequences.

Analyzing the FPR and FNR, it is evident that models like LSTM and Logistic Regression manage to maintain low rates in both positive and negative cases. This is further supported by their TPR and TNR values, showcasing their ability to accurately identify true cases. Finally, the Detection Error Tradeoff (DET) curve reveals that models with higher ROC values, such as MLP and Random Forest, have lower DET curves, while Naive Bayes, with its lower AUC, exhibits a higher DET curve. This suggests that models with better ROC performance are more capable of minimizing detection errors. This relationship underscores the importance of choosing models with strong performance across all evaluation metrics, as they are likely to exhibit better overall classification accuracy and reliability.

\begin{figure}
    \centering
    \includegraphics[width=\linewidth]{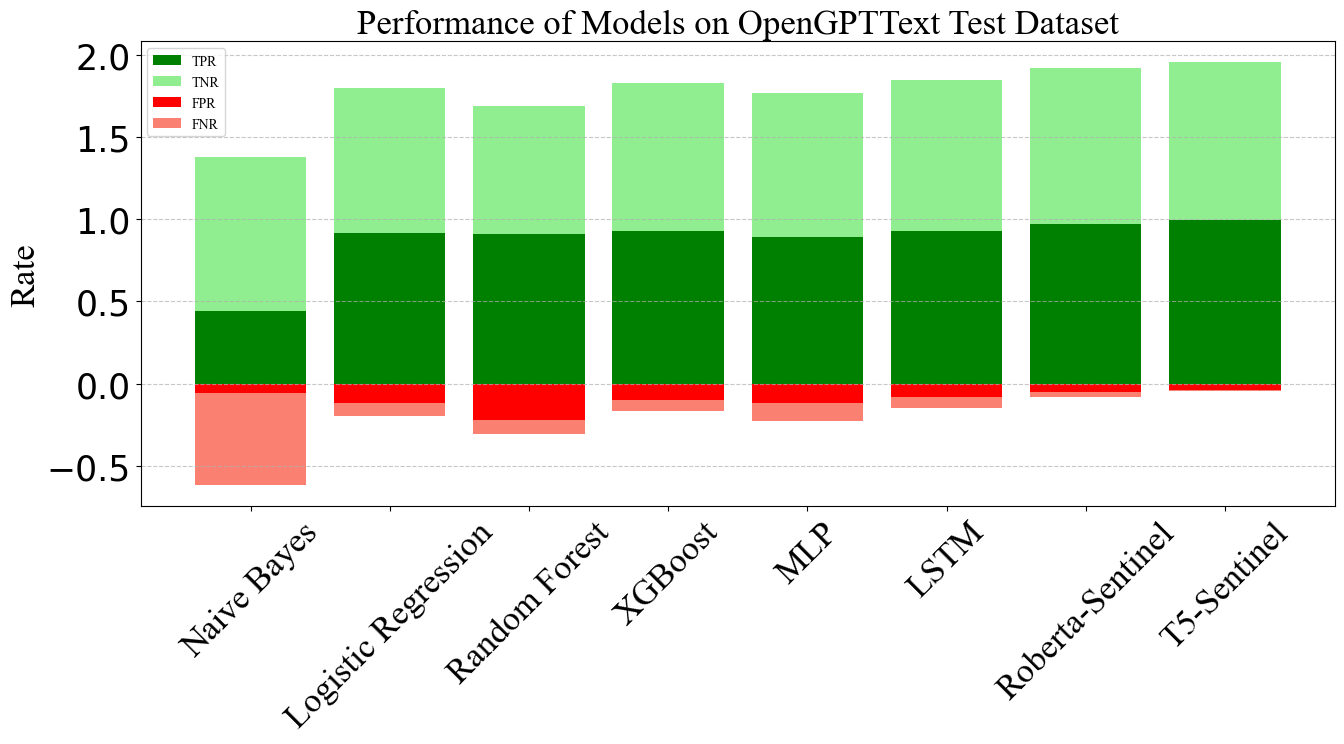}
    \caption{The stacked bar graph compares the performance of eight models on the OpenGPTText-Final Dataset in distinguishing AI-generated and human-written text. The T5-Sentinel model shows the highest effectiveness with a 0.996 probability of correctly identifying AI-generated text (TPR) and 0.96 for human-written text (TNR). Roberta-Sentinel follows closely with a TPR of 0.97 and TNR of 0.95. In contrast, Naive Bayes has a 0.94 TNR but only a 0.44 TPR, favoring human-written text detection. Random Forest struggles with a lower TNR of 0.78, while Logistic Regression and LSTM perform consistently around 0.88-0.93 for both TPR and TNR. Overall, T5-Sentinel provides the most balanced and accurate classification for this dataset.}

    \label{fig:enter-label}
\end{figure}
\begin{table}[h]
    \centering
    \begin{tabular}{|c|c|c|c|c|c|c|}
        \hline
        \textbf{Model} & \textbf{Accuracy} & \textbf{F1-score} & \textbf{FPR} & \textbf{FNR} & \textbf{TNR} & \textbf{TPR} \\
        \hline
        Naive Bayes & 0.69 & 0.59 & 0.06 & 0.56 & 0.94 & 0.44\\
        \hline
        Logistic Regression & 0.90 & 0.91 & 0.12 & 0.08 &0.88 & 0.92\\
        \hline
        Random Forests & 0.85 & 0.86 & 0.22 & 0.09 & 0.78 & 0.91\\
        \hline
        XGBoost & 0.91 & 0.92 & 0.10 & 0.07 & 0.90 & 0.93\\
        \hline
        MLP & 0.89 & 0.89 & 0.12 & 0.11 & 0.88 & 0.89 \\
        \hline
        LSTM & 0.92 & 0.92 & 0.08 & 0.07 & 0.92 & 0.93\\
        \hline
        RoBERTa & 0.95 & 0.95 & 0.05 & 0.03 & 0.95 & 0.97\\
        \hline
        T5 & 0.98 & 0.97 & 0.04 & 0.004 & 0.96 & 0.996\\
        \hline
    \end{tabular}
    
    \caption{Evaluation Results on OpenGPTText Dataset}
    \label{tab:evaluation}
\end{table}

\subsection{RoBERTa-Sentinel and T5-Sentinel}

In the analysis of the RoBERTa-Sentinel and T5-Sentinel models applied to the OpenGPTText-Final Dataset, both models demonstrated excellent performance across various evaluation metrics. The RoBERTa-Sentinel model shows a False Positive Rate (FPR) of 0.05 and a False Negative Rate (FNR) of 0.03, with a True Negative Rate (TNR) of 0.95 and a True Positive Rate (TPR) of 0.97. These values indicate that RoBERTa-Sentinel is highly effective in correctly classifying both positive and negative instances, with minimal errors in misclassifications. This strong performance is further illustrated in Table 4, and the detailed classification breakdown can be visualized in Figure 3.

The T5-Sentinel model performs even better, with an FPR of 0.04 and a significantly lower FNR of 0.004. Its TNR stands at 0.96, and its TPR reaches 0.996, reflecting near-perfect accuracy in distinguishing between AI-generated and human-written text. The exceptionally low FNR suggests that T5-Sentinel is particularly adept at correctly identifying positive instances, which in this context could be AI-generated text. The high TPR and TNR further highlight its precision and reliability in classification tasks.

When comparing these models' ROC values, RoBERTa-Sentinel achieves an impressive ROC of 0.99, indicating its strong performance across various threshold settings. However, T5-Sentinel surpasses this with a perfect ROC of 1.00, signifying flawless discrimination between classes without any overlap or error.

In terms of F1-score and accuracy, T5-Sentinel outshines RoBERTa-Sentinel, showcasing its remarkable ability to strike a balance between precision and recall. This performance can be largely attributed to the unique architecture of T5. As a sequence-to-sequence (seq2seq) model, T5 was initially designed for text generation tasks\cite{raffel2023exploringlimitstransferlearning}, which involve handling both input and output sequences. This capability allows T5 to excel at recognizing complex patterns, such as differentiating between AI-generated and human-written text. The seq2seq framework\cite{wang-etal-2021-codet5} gives T5 the edge in capturing subtle relationships within the text, making it particularly adept at this kind of classification task.

However, while RoBERTa-Sentinel and T5-Sentinel represent excellent performance, traditional machine learning models like Logistic Regression, Random Forest, and LSTM, MLP should not be ignored. These models still demonstrate considerable potential. For instance, LSTM’s ability to handle sequential data, Random Forest’s ensemble strength, and Logistic Regression’s robustness in linear scenarios show that traditional models can still deliver strong performance. Their ability to strike a balance between complexity and efficiency makes them viable options in human-written or AI-generated text classification tasks.

Both RoBERTa-Sentinel and T5-Sentinel deliver outstanding results, with T5-Sentinel leading in most metrics. These models are setting new standards in distinguishing AI-generated text from human-written content, demonstrating the power of advanced deep learning techniques. However, traditional models should not be overlooked. They continue to offer dependable and understandable solutions in these classification tasks. This highlights the importance of a balanced approach one that values the strengths of both modern and established methods when selecting the best tool for detecting whether text is AI-generated or human-written.

\subsection{Testing on custom dataset}
In evaluating our custom test dataset, T5-Sentinel and RoBERTa-Sentinel have demonstrated exceptional performance, showcasing the strengths of complex deep learning techniques. T5-Sentinel, with its seq2seq architecture, achieved an accuracy of 0.88 and an F1-score of 0.88. It excels in balancing precision and recall, evidenced by its false negative rate of 0.08 and false positive rate of 0.16. RoBERTa-Sentinel also performed strongly, attaining an accuracy of 0.92 and an F1-score of 0.92, with a low false negative rate of 0.04 and a high true positive rate of 0.96. MLP, being one of the traditional ML Model matched this level of performance with an accuracy of 0.88 and an F1-score of 0.89, demonstrating its capability to handle complex text patterns effectively.

\begin{figure}
    \centering
    \includegraphics[width=\linewidth]{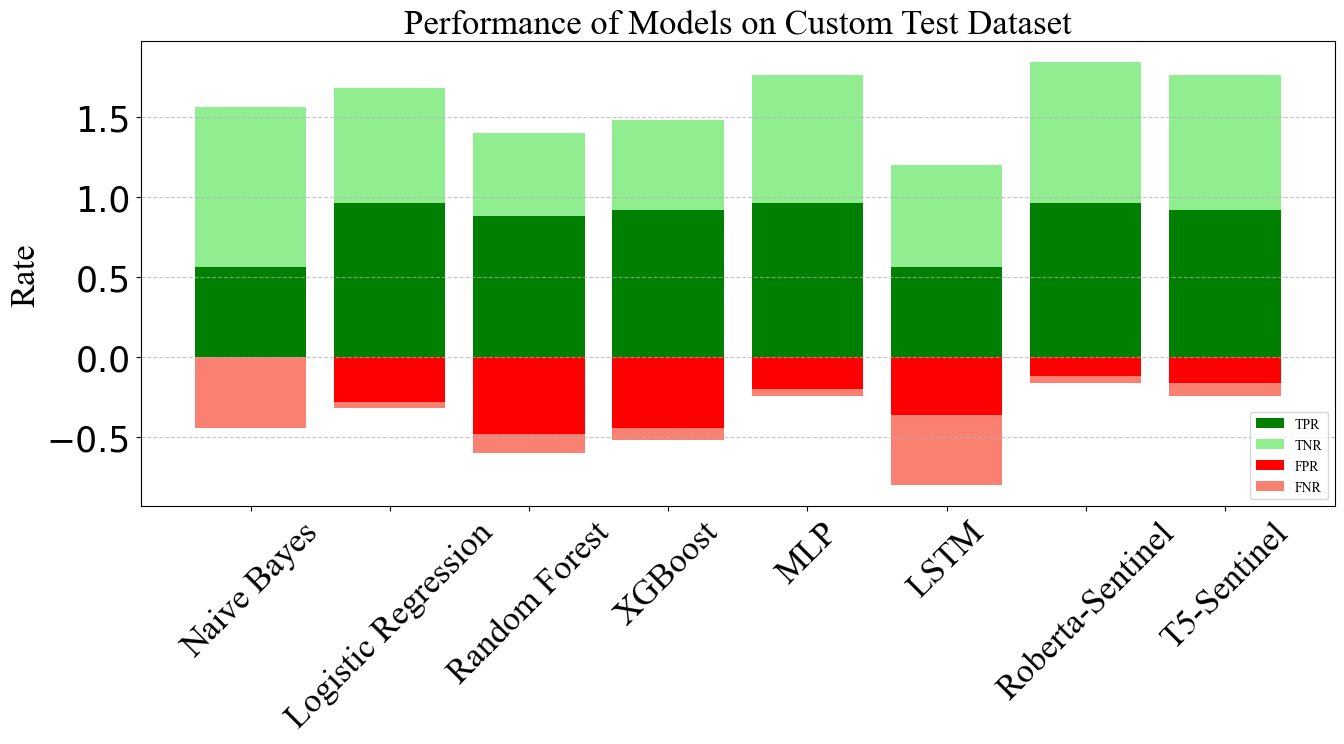}
    \caption{The stacked bar graph compares eight models' performance in classifying AI-generated and human-written text based on True Positive Rate (TPR), True Negative Rate (TNR), False Positive Rate (FPR), and False Negative Rate (FNR). RoBERTa excels with a TPR of 0.96 and a TNR of 0.88, demonstrating strong performance in identifying both text types while maintaining a low FPR of 0.12 and FNR of 0.04. MLP also performs well with a TPR of 0.96 and a TNR of 0.80, though it has a slightly higher FPR of 0.20. Logistic Regression achieves a TPR of 0.96 and a TNR of 0.72, with an FPR of 0.28. T5 shows balanced performance with a TPR of 0.92 and a TNR of 0.84, alongside an FPR of 0.16. Naive Bayes has the highest TNR of 1.00 but a lower TPR of 0.56 and FNR of 0.44, indicating challenges in detecting AI-generated text. Random Forests and XGBoost show moderate results, while LSTM has the lowest TPR of 0.56 and TNR of 0.64, suggesting less effective performance in distinguishing between text types.}
    \label{fig:enter-label}
\end{figure}
Other Traditional machine learning models also offer strong competition. Logistic Regression, with an accuracy of 0.84 and an F1-score of 0.86, shows effective performance with a notably low false negative rate of 0.04. XGBoost, achieving an accuracy of 0.72 and an F1-score of 0.78, maintains a false negative rate of 0.08. Although LSTM has lower accuracy and F1-score compared to the advanced models, it remains relevant due to its ability to capture sequential dependencies.

Models such as Naive Bayes and Random Forests, while performing lower compared to T5-Sentinel, RoBERTa-Sentinel, and MLP, still contribute meaningfully. Naive Bayes achieved an accuracy of 0.78 and an F1-score of 0.72, with a higher false positive rate, while Random Forests had an accuracy of 0.70 and an F1-score of 0.75.

.

\begin{table}[h]
    \centering
    \begin{tabular}{|c|c|c|c|c|c|c|}
        \hline
        \textbf{Model} & \textbf{Accuracy} & \textbf{F1-Score} & \textbf{FPR} & \textbf{FNR} & \textbf{TNR} & \textbf{TPR}\\
        \hline
        Naive Bayes & 0.78 & 0.72 & 0 & 0.44 & 1 & 0.56\\
        \hline
        Logistic Regression & 0.84 & 0.86 & 0.28 & 0.04 & 0.72 & 0.96\\
        \hline
        Random Forests & 0.70 & 0.75 & 0.48 & 0.12 &0.52 & 0.88\\
        \hline
        XGBoost & 0.72 & 0.78 & 0.44 & 0.08 & 0.56 & 0.92\\
        \hline
        MLP &  0.88 & 0.89 & 0.2 & 0.04 & 0.8 & 0.96\\
        \hline
        LSTM & 0.60 & 0.60 & 0.36 & 0.44 & 0.64 &  0.56\\
        \hline
        RoBERTa & 0.92 & 0.92 & 0.12 & 0.04 & 0.88 & 0.96\\
        \hline
        T5 & 0.88 & 0.88 & 0.16 & 0.08& 0.84 & 0.92\\
        \hline
    \end{tabular}
    \caption{Evaluation Results on Custom Test Dataset}
    \label{tab:evaluation}
\end{table}
Overall, while T5-Sentinel, RoBERTa-Sentinel, and MLP set high benchmarks in model performance, traditional models like Logistic Regression and XGBoost offer strong competition and practical value. For detailed performance metrics and further comparative information, please refer to Table \ref{tab:evaluation} and Figure 4.
\section{Model Explainability with LIME}
\subsection{Overview of LIME}

Interpreting complex machine learning models is essential, especially as these models increasingly impact critical decisions. Many studies focus primarily on achieving high accuracy in distinguishing human-written text from AI-generated content but often neglect the need to understand how these models derive their predictions. The Local Interpretable Model-agnostic Explanations (LIME) method addresses this by providing a practical approach to making sense of model outputs. LIME works by approximating the model’s behavior around specific predictions and highlighting the features that shape its results. Our research used LIME to gain a clearer understanding of the model’s reasoning, emphasizing the importance of interpretability in AI research.

\subsection{Application of LIME Across Models}
LIME has been effectively used with a variety of machine learning models, ranging from basic models like those using TF-IDF vectorization to more complex ones such as Long Short-Term Memory (LSTM) networks, RoBERTa, and T5 transformers. For traditional models, LIME assesses the impact of specific features, like words or phrases, on the final prediction. For example, with a linear classifier trained on TF-IDF features, LIME might pinpoint important terms that play a significant role in the classification result.

In deep learning models like LSTMs, LIME helps explain how sequences of words affect the model’s output. LSTMs need preprocessing steps such as tokenization and padding to manage variable-length sequences. By using LIME, you can identify which tokens or word sequences had a major effect on the prediction, giving a better view of how the model works internally.

For transformer models like RoBERTa and T5, which use attention mechanisms, LIME breaks down the influence of individual tokens or phrases. These models also require specific preprocessing, such as tokenization with attention masks, to properly format the input data. LIME shows how certain tokens, which might represent key semantic elements in the text, influence the model’s predictions.

\subsection{Preprocessing steps}
Before applying LIME, data preprocessing steps vary depending on the model type. For LSTM models, text data is tokenized and padded to ensure consistent input length across samples, which is necessary for effective training. In traditional models using TF-IDF vectorization, the text is transformed into a numerical feature space where the importance of words is weighted based on their frequency in the corpus relative to their rarity. Transformer models, such as RoBERTa and T5, require tokenization and the addition of attention masks to ensure that the model can process the input text in a way that captures its contextual relationships.

\subsection{Interpreting Feature Importance}
LIME provides clarity on which features, such as specific words or tokens, have the most impact on a model’s predictions. By identifying these key features, LIME reveals what drives the model’s output.

Understanding the importance of these features can uncover broader patterns within the data. For instance, if certain words or phrases consistently emerge as significant, it might indicate recurring themes or tendencies in the model’s responses. Recognizing these patterns helps in refining the model by adjusting its focus or correcting any imbalances in how features are valued.

Additionally, this understanding can guide improvements in the model. If certain features are found to dominate the prediction process, modifications can be made to either reduce their influence or enhance the model’s ability to handle a diverse range of inputs effectively. This approach not only boosts the model’s performance but also ensures its reliability and fairness in various applications..

\subsection{Analyzing Feature Contributions for across different Models for Different Samples}

We perform LIME analysis on samples from the \texttt{OpenGPTText-Final Dataset} and \texttt{Custom Test Dataset}, including newly generated samples that the models had never encountered before. The goal is to understand the rationale behind the model's classification decisions specifically, why certain samples were labeled as AI-generated or human-written. We applied LIME across various models, ranging from Naive Bayes and Logistic Regression to advanced architectures like T5-Sentinel, and identified the top 10 features contributing to each model's predictions. This approach provided valuable clarity into the decision-making processes of these deep learning models, which are often regarded as black boxes.

Let's now focus on a specific example generated by \texttt{Chat-GPT 4o} and analyzed using LIME across different models. Refer to Figure 5 for the sample text, which is presented in a condensed format. Additional details can be found in the appendix.
\begin{figure}
    \centering
    \includegraphics[width=0.8\textwidth]{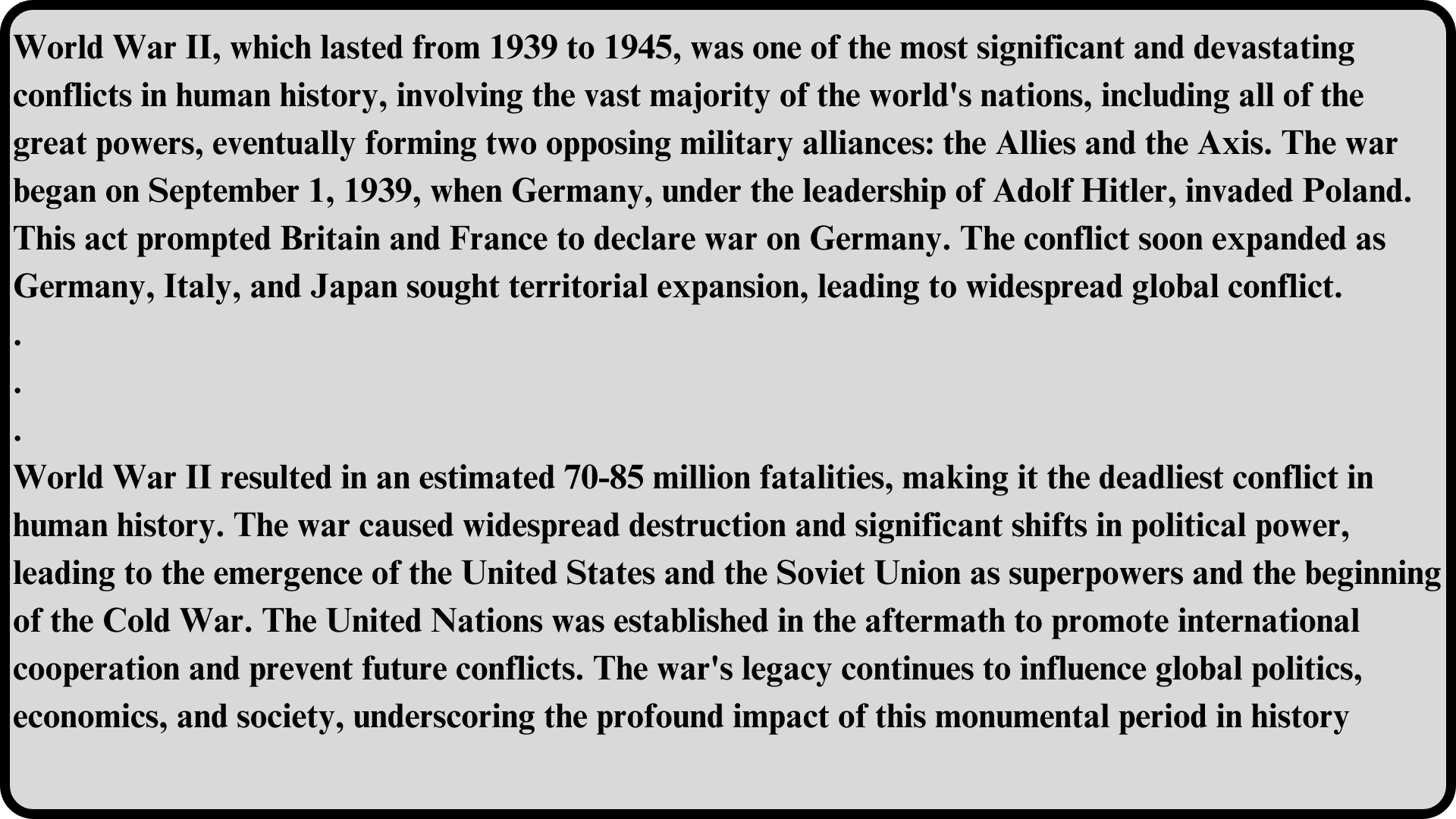}
    \caption{This is the one of the sample generated by ChatGPT-4o, detailed sample can be accessed in the Github repository link provided in Section 6.}
    \label{fig:enter-label}
\end{figure}
All the models tested, from Naive Bayes to T5-Sentinel, classified the sample as AI-generated, albeit with varying probabilities. Naive Bayes, for instance, gave the lowest confidence score of 0.54 for AI generation, with \texttt{significant} being the top contributing feature for the AI classification. Conversely, the word \texttt{war} was the leading feature contributing to the classification of the text as human-written, with a probability of 0.46.

The more advanced models, such as T5-Sentinel, MLP, and LSTM, were among the top performers, predicting the sample as AI-generated with probabilities of 1.0, 1.0, and 0.93, respectively. Interestingly, \texttt{significant} was consistently identified as a key feature in most models, contributing 0.29 for LSTM and 0.11 for MLP, while T5-Sentinel had \texttt{was} as its top contributing factor. This observation suggests that \texttt{significant}, a frequently occurring word in the text, plays a crucial role in the classification decision across multiple models, likely due to its frequent usage in the paragraph, which may be a stylistic marker associated with AI-generated content.

During this analysis, \texttt{the} emerged as one of the top features contributing to the classification of some AI-generated samples. This finding prompted a discussion about whether to remove such common words in preprocessing. However, upon reflecting on human writing styles, we observed that non-native speakers and individuals less concerned with formal writing might not consistently use articles like \texttt{the,an} and \texttt{a}. In contrast, large language models (LLMs), trained on extensive datasets with attention mechanisms, are more likely to use these words accurately. Therefore, we decided to retain these words in our study to preserve the authenticity of the analysis.

Overall, this LIME-based study provided us with a deeper understanding of the patterns or stylistic elements that each model learns and uses for predictions. The use of explainable AI techniques significantly enhanced the transparency and interpretability of our analysis, offering clear insights into the complex decision-making processes of deep learning models.

\subsection{Summary and Implications}
LIME is a valuable tool for making complex machine learning models easier to understand. It provides clear explanations for individual predictions by showing which features are influencing the model's decisions. This helps us grasp how models work and builds trust in their predictions, regardless of whether the models are traditional or more modern.

In our task of distinguishing between human and AI-generated text, LIME is especially helpful. It reveals which factors the model is considering when making its decisions. For instance, LIME can show if the model is focusing on certain language patterns typical of human or AI writing. This transparency ensures that the model's decisions are not just accurate but also reasonable. By clarifying how the model operates, LIME helps us validate its performance and boosts our confidence in its ability to differentiate between human and AI-generated content.

\begin{figure}
    \centering
    \includegraphics[width=\linewidth]{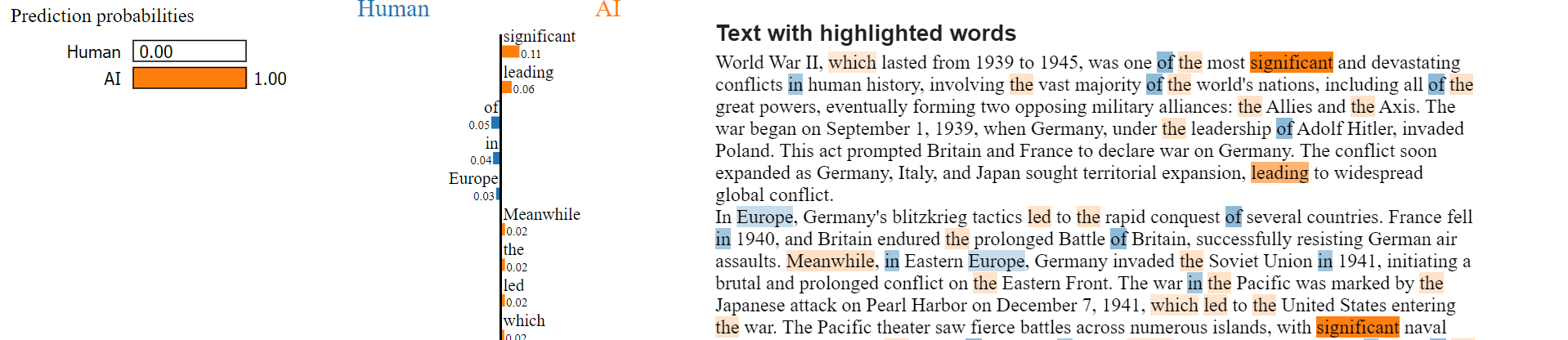}
    \caption{
This LIME plot for the MLP model illustrates the significance of various features in classifying the sample as AI.We can observe from the figure 6 that the sample text given in figure 5 is  predicted as AI generated with probability score of 1 where \texttt{significant, leading} are the top 2 contributing features for the prediction with the score of 0.31 and 0.06 respectively}

    \label{fig:enter-label}
\end{figure}

\begin{figure}
    \centering
    \includegraphics[width=\linewidth]{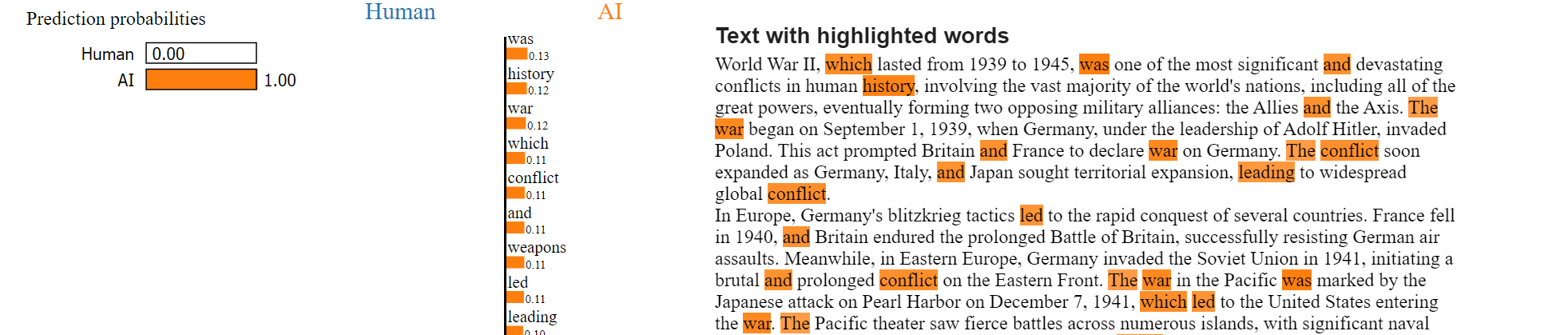}
    \caption{
This LIME plot for the T5-Sentinel model illustrates the significance of various features in classifying the sample as AI.We can observe from the figure 7 that the sample text given in figure 5 is  predicted as AI generated with probability score of 1 where \texttt{was, history} are the top 2 contributing features for the prediction with the score of 0.13 and 0.12 respectively}

    \label{fig:enter-label}
\end{figure}

\begin{figure}
    \centering
    \includegraphics[width=\linewidth]{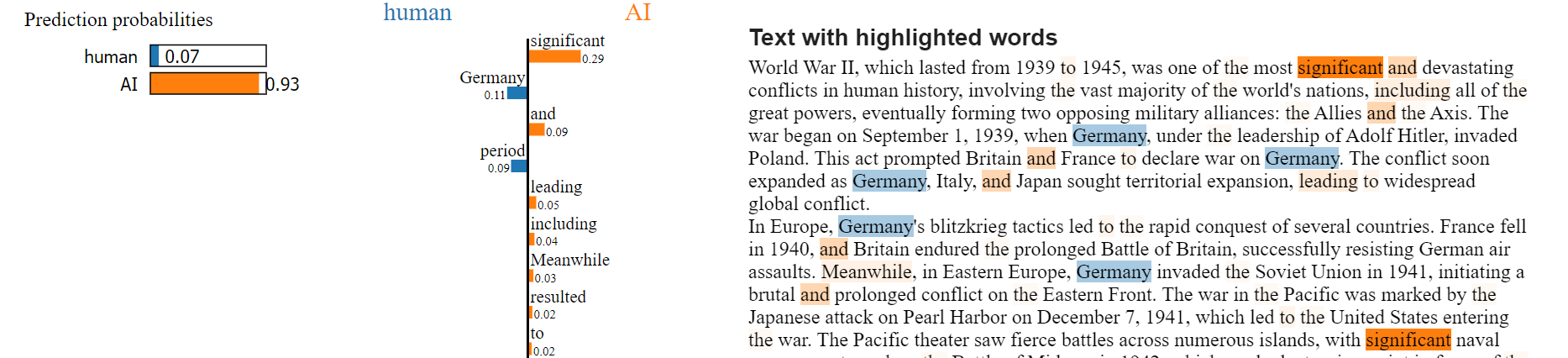}
    \caption{
This LIME plot for the LSTM model illustrates the significance of various features in classifying the sample as AI.We can observe from the figure 8 that the sample text given in figure 5 is  predicted as AI generated with probability score of 0.93 and Human written  with probability score of 0.07 where \texttt{significant} is the topmost contributing feature for AI generated with score of 0.29 and \texttt{Germany} is topmost contributing feature for Human written text with the score of 0.11}

    \label{fig:enter-label}
\end{figure}

\section{Conclusion and Discussion}
This study explored the effectiveness of both traditional machine learning models and advanced deep learning models in distinguishing AI-generated text from human-written content. Notably, while deep learning models like T5 and RoBERTa achieved high accuracy rates of 0.98 and 0.95 on the OpenGPT dataset, traditional models like  XGBoost and Random Forests also performed commendably. XGBoost, for instance, achieved an impressive accuracy of 0.91, closely matching the performance of more complex models. These results highlight that traditional models, despite their simplicity, remain powerful tools for text classification tasks, particularly when considering their interpretability and efficiency.

On a custom test dataset spanning various domains, traditional models continued to demonstrate their robustness. MLP maintained a strong performance with an accuracy of 0.90, proving its versatility across different text types. While advanced models like RoBERTa excelled with 0.92 accuracy, the consistent performance of traditional models highlights their reliability and potential for use in real-world applications, where computational resources and model explanability are crucial.

Finally, we employed LIME (Local Interpretable Model-agnostic Explanations) analysis on both the OpenGPTText-final dataset and a custom test dataset, which included previously unseen samples. This approach allowed us to uncover the key features driving the models' decisions in distinguishing AI-generated text from human-written content. Unlike many studies in this field, our use of LIME provided a deeper understanding of the reasoning behind each model's classification, revealing distinct patterns in feature importance across different models. This analysis underscored the value of interpretability, showing that even in complex deep learning models, we can gain meaningful insights into how decisions are made, thereby enhancing the transparency and reliability of AI systems.

While our results are promising, there are limitations to this study. One major limitation is the size and scope of the datasets used. Although the OpenGPT and Custom Test Datasets provided a broad range of text samples, a larger and more diverse dataset could further validate the findings and enhance the generalizability of the models. Additionally, the study predominantly focused on ChatGPT-generated text, which limits the applicability of our conclusions to other LLMs. The binary classification approach may also fail to capture nuances in text generation that could be better captured through more sophisticated methods.

Future research will aim to address these limitations by expanding the dataset to include a wider variety of text sources and LLMs. We also plan to explore the integration of traditional models with Sophisticated deep learning architectures to create a hybrid model that leverages the strengths of both approaches. This proposed model will utilize a majority voting mechanism among traditional models to improve the robustness of predictions. Moreover, we intend to apply SHAP analysis to the dataset, providing more granular details into model decision-making and enhancing the explainability of the results. Extending this study to include other LLMs will also be a priority, aiming to demonstrate that traditional models can perform on par with Sophisticated deep learning methods across diverse contexts.

In summary, our study highlights the continued importance of traditional machine learning models in the rapidly evolving field of text classification. While modern deep learning models like RoBERTa-Sentinel and T5-Sentinel have pushed the boundaries of accuracy, traditional models still hold their own, especially in settings where resources are limited. By incorporating explainability techniques like LIME, we were able to gain deeper revelations into how these models make decisions, reinforcing the value of transparency in AI. Looking ahead, blending the strengths of both traditional and modern approaches could lead to more balanced, effective, and interpretable AI systems.

\section{Dataset and code availability}
The dataset and the entire code has been uploaded at this Github repository: 

\texttt{https://github.com/pdjoshi-30/HULLMI-HUMAN-VS.-LLM-IDENTIFICATION-WITH-EXPLAINABILITY}

\bibliographystyle{unsrt}  
\bibliography{references}  %%% Remove comment to use the external .bib file (using bibtex).

\end{document}